\RequirePackage{fix-cm}
\documentclass[twocolumn]{svjour3}          % twocolumn

\smartqed  % flush right qed marks, e.g. at end of proof
\usepackage{graphicx}

\usepackage{multirow}
\usepackage{amsmath,amssymb}
\usepackage[switch,pagewise]{lineno}
\newcommand*\patchAmsMathEnvironmentForLineno[1]{
    \expandafter\let\csname old#1\expandafter\endcsname\csname #1\endcsname
    \expandafter\let\csname oldend#1\expandafter\endcsname\csname end#1\endcsname
    \renewenvironment{#1}
    {\linenomath\csname old#1\endcsname}
    {\csname oldend#1\endcsname\endlinenomath}}
\newcommand*\patchBothAmsMathEnvironmentsForLineno[1]{
    \patchAmsMathEnvironmentForLineno{#1}
    \patchAmsMathEnvironmentForLineno{#1*}}
\AtBeginDocument{
\patchBothAmsMathEnvironmentsForLineno{equation}
\patchBothAmsMathEnvironmentsForLineno{align}
\patchBothAmsMathEnvironmentsForLineno{flalign}
\patchBothAmsMathEnvironmentsForLineno{alignat}
\patchBothAmsMathEnvironmentsForLineno{gather}
\patchBothAmsMathEnvironmentsForLineno{multline}
}
\begin{document}

\title{Modeling Task Immersion based on Goal Activation Mechanism%\thanks{Grants or other notes
%about the article that should go on the front page should be
%placed here. General acknowledgments should be placed at the end of the article.}
}
% \subtitle{Do you have a subtitle?\\ If so, write it here}

%\titlerunning{Short form of title}        % if too long for running head

\author{Kazuma Nagashima$^{1}$            \and
        Jumpei Nishikawa$^{1}$        \and
        Junya Morita$^{1,2}$
}

%\authorrunning{Short form of author list} % if too long for running head

\institute{
Kazuma Nagashima \\
\email{nagashima.kazuma.16@shizuoka.ac.jp} \\
Junya Morita \\
\email{j-morita@inf.shizuoka.ac.jp} \\ \at 
$^{1}$ Department of Information Science and Technology, Graduate School of Science and Technology, Shizuoka University
              3-5-1 Johoku, Chuo-ku, Hamamatsu-shi, Shizuoka-ken, 432-8011, Japan \and 
$^{2}$              Department of Behavior Informatics, Faculty of Informatics, Shizuoka University, 3-5-1 Johoku, Chuo-ku, Hamamatsu-shi, Shizuoka-ken, 432-8011, Japan
              % \and
}           
              % 3-5-1 Johoku, Chuo-ku, Hamamatsu-shi, Shizuoka-ken, 432-8011, Japan \\
              % Tel.: +123-45-678910\\
              % Fax: +123-45-678910\\
              % \email{nagashima.kazuma.16@shizuoka.ac.jp}           %  \\
%             \emph{Present address:} of F. Author  %  if needed
           % \and
           % S. Author \at
           %    second address

% \institute{F. Author \at
%               first address \\
%               Tel.: +123-45-678910\\
%               Fax: +123-45-678910\\
%               \email{fauthor@example.com}           %  \\
% %             \emph{Present address:} of F. Author  %  if needed
%            \and
%            S. Author \at
%               second address
% }

\date{Received: date / Accepted: date}
% The correct dates will be entered by the editor

\newpage
\maketitle

\begin{abstract}
Immersion in a task is a prerequisite for creativity. However, excessive arousal in a single task has drawbacks, such as overlooking events outside of the task. To examine such a negative aspect, this study constructs a computational model of arousal dynamics where the excessively increased arousal makes the task transition difficult. The model was developed using functions integrated into the cognitive architecture Adaptive Control of Thought-Rational (ACT-R). Under the framework, arousal is treated as a coefficient affecting the overall activation level in the model. In our simulations, we set up two conditions demanding low and high arousal, trying to replicate corresponding human experiments. In each simulation condition, two sets of ACT-R parameters were assumed from the different interpretations of the human experimental settings. The results showed consistency of behavior between humans and models both in the two different simulation settings. This result suggests the validity of our assumptions and has implications of controlling arousal in our daily life.
% \keywords{First keyword \and Second keyword \and More}
% \PACS{PACS code1 \and PACS code2 \and more}
% \subclass{MSC code1 \and MSC code2 \and more}

\end{abstract}
\section{Introduction} \label{introductin}
Concentrating on the task at hand, which sometimes relates to psychological states called {\it immersion} or {\it flow}\footnote{
We use ``immersion'' as a synonym of ``flow'' in terms of Csikszentmihalyi's flow theory \cite{Csikszentmihalyi1990}. In this theory, high arousal is pre-requisite of the state of flow.}, is widely acknowledged as a precursor to achieving high performance in the task \cite{Csikszentmihalyi1990}. During such states, intensified concentration, coupled with a feeling of absolute control over the task, can foster creative outcomes, albeit sometimes at the cost of losing self-consciousness.

This paper endeavors to construct a computational model elucidating the psychophysiological dynamics underlying an immersive state. Our model places particular emphasis on arousal, defined as ``{\it a state of physiological activation or cortical responsiveness, associated with sensory stimulation and activation of fibers from the reticular activating system} \cite{vandenbos2007apa}.''  From this definition, we understand arousal as {\it activated} brain states, which also relate to {\it attention}. One famous theory on the relation between arousal (activation) and attention is the cue utilization theory \cite{easterbrook1959effect}, which points out that intensified arousal narrows attention to the external environment.

In contrast to simple explanations of concentration based solely on the level of attention, those based on arousal connect to fundamental factors related to emotion and physiology (e.g., \cite{russell1980circumplex}). Regarding such factors, the existence of an optimal level of arousal has been pointed out in several theories \cite{Csikszentmihalyi1990,oxendine1970emotional,yerkes1908relation}, which suggest that both excessively high and exceedingly low arousal levels compared to the optimal level have detrimental effects on task motivation. In essence, task motivation is related with arousal level through an inverted U-shaped function, with the peak determined by task difficulty. Difficult tasks require high arousal levels. However, if tasks are too challenging for low-aroused workers, they lose motivation due to factors such as anxiety. Conversely, in simpler tasks where elevated arousal isn't needed, highly aroused workers experience a loss of motivation due to boredom.

Mismatching the level of arousal impacts task performance, not only motivation. In situations of low arousal, boredom with the task can lead to reduced performance due to distractions, a phenomenon often associated with mind-wandering \cite{Feng2013mind}. Conversely, decreased performance caused by high arousal can occur in difficult tasks, where a certain level of concentration is required. As noted earlier, highly elevated arousal leads a deficit in attention allocation \cite{easterbrook1959effect}. Thus, in a situation requiring high arousal, workers often overlook information outside of their current focus. % are usually engage in several tasks multitasking must distribute their attention properly among several tasks (goals) based on the difficulty and importance of them. However, when someone becomes deeply immersed in one task, their ability to allocate attention to other tasks is compromised, leading to the overlooking of information that is important for other tasks. 
Consequently, overall performance in situations involving multiple goals, such as multitasking, declines when workers' arousal exceeds the optimal level for the task at hand \cite{diamond2007temporal}\footnote{Difficult tasks require more arousal than easier tasks. However, it is not true that higher arousal is always appropriate for the difficult task.}.

Numerous studies have explored human cognitive functions in relation to arousal levels. However, no models have integrated the cognitive functions that produce the temporal dynamics linking arousal changes to performance outcomes. We believe that this temporal integration of arousal and performance can be achieved through the utilization of a cognitive architecture that combines the fundamental cognitive functions involved in diverse tasks. 

In this study, we depict this process through the utilization of a cognitive architecture known as ACT-R (Adaptive Control of Thought-Rational \cite{Anderson:2007}). Similar to numerous other cognitive architectures, ACT-R offers modules that mirror brain functions employed consistently across multiple tasks (see \cite{Kotseruba2020} as a comprehensive review). The functions integrated into ACT-R modules encompass visual processing, motor actions, goal management, memory storage, and procedural memory. In addition to those symbolic modules, ACT-R includes components called subsymbolic process, which closely relates to motivation and emotional process. This study represents the process of immersion at a situation where multitasking (managing multiple goals) is required by combining these modules.

In the upcoming section, we will explore relevant studies pertinent to the aforementioned objective of this research. Subsequently, we will delve into the specific human behaviors related to arousal in a multitasking context. Following this, we will elaborate on the ACT-R model, which integrates various foundational cognitive components to simulate these specific behavioral patterns. The simulation results revealed a sustained decrease of response times to the subgoal under a condition where low arousal is required, while in a condition where high arousal is needed, response times to the subgoal remained consistently high. In the final section, we will thoroughly examine the implications and limitations of this study.

\section{Related Works}
This study combines primitive functions in ACT-R to explain arousal dynamics leading to deficits in attention allocation. This section presents general surveys on computational models of physiological mechanism and surveys focusing on research using ACT-R to explain arousal dynamics and multitasking.

\subsection{Computational Models of Biological Homeostasis}\label{homeostasis}
Physiological processes drive human arousal. Therefore, the optimal level of arousal described in the previous section can be interpreted in the context of homeostasis maintenance, which is a general self-regulating process that maintains its optimal values of physiological variables \cite{billman2020homeostasis,cannon1929organization}. According to this theory, organisms can adapt to changing environments because of homeostasis.

Computationally, homeostasis has been explained by the theory of predictive coding \cite{clark2013whatever,friston2010free,rao1999predictive,SETH2013565}, which posits that organisms desire to minimize long-term mismatches (errors, surprises) between predictions from experience and perceptions of the external environment. 
There are several ways to reduce these discrepancies. One way is to master the task so that the prediction is accurate and performance is close to the predicted states. However, mastering a particular task also induces mismatches in the future environment. 
Thus, the theory describes human behavior as a balance between minimizing prediction error for the current task (i.e., exploitation) and increasing prediction accuracy in future environments (i.e., exploration) \cite{schwartenbeck2013exploration}. Such a balancing behavior links to the optimal arousal theory by directly connecting surprises generated by prediction errors to arousal potential \cite{yanagisawa2021free}.  

The relationship between prediction error and arousal (activation) is also explained by the Expected Value of Control (EVC) theory \cite{SHENHAV2013217}. This theory explains the function of the dorsal anterior cingulate cortex (dACC) as the coordinator of human behavior by minimizing the difference between expected and actual rewards (reward prediction error). These coordination functions are instantiated as an EVC value, which is computed from two components: ``identity,'' which is a control signal distinguishing several goals, and ``intensity,'' which is activation of neurons relating to each control signal. The function is balanced by an expected reward and cost polynomial, and the EVC value draws an inverse U-shape when ``intensity'' is manipulated. Organisms are considered to choose the behavior with the highest EVC value.

So far, we have introduced theories explaining the inverted U-shaped relationship between task motivation/performance and arousl levels. A fundamental principle across these theories is the self-regulation mechanism of arousal. The continuation of the same task leads to the saturation of prediction errors, increasing the desire to explore new environments \cite{schwartenbeck2013exploration}. However, the above theories has difficulty describing the detailed and concrete process of arousal changing over time. To solve this problem, we need a process model in which several components are interacted.

\subsection{ACT-R Models Regarding Arousal Dynamics}
As noted in Section \ref{introductin}, this study uses ACT-R to model cognitive processes relating to arousal dynamics. Several researchers have developed ACT-R models explaining how inappropriate arousal levels affect task motivation/performance. For example, Vugt et al. \cite{van2015modeling} developed a model of mind-wandering using the activation mechanism of ACT-R. In their model, mind-wandering is represented as memory recalls of task-unrelated contents. Through the execution of the monotonous task, the activation of these memory contents is gradually strengthened, eventually leading to mind-wandering and performance decline.

Other studies have focused on fatigue, a psychophysiological state associated with low arousal. Gunzelmann et al. \cite{gunzelmann2009using} constructed a model representing the effects of fatigue on the execution of a continuous perceptual-motor coordination task. Speciﬁcally, they manipulated the parameter relating to the computations of selection probability of specific action to execute the task. Gunzelmann et al. \cite{gunzelmann2012diminished} also constructed a mechanism for fatigue in memory activation, which affects the success of memory retrieval during the task. These changes in numerical parameters of ACT-R can define an inverse U-shaped curve representing the relation between the task continuation and performance measured by reaction time \cite{atashfeshan2017determination} \footnote{Although this inverse U-shape relates to optimal arousal, these two are not the same because the influence of task continuation is not limited to arousal level.}.

Furthermore, another ACT-R model explains motivation related to arousal using the EVC theory presented in the previous section. Yang and Stocco \cite{yang2022iccm} represents ``identity'' in EVC theory as the goal contents of ACT-R and ``intensity'' as activation of the goal. In their study, the U-shaped curve of the optimal arousal level was reproduced when several conflict goals existed.

However, these studies have not explicitly discussed the correspondence of these parameter changes to human physiological mechanisms. Concerning the logic behind these mechanisms, Ritter \cite{ritter2009two} defined emotion as physiological substrates affecting cognitive parameters. This idea has been instantiated in ACT-R/$\Phi$ \cite{Dancy:cn-BYwiN}, which combines cognitive processes in ACT-R with physiological mechanisms. Although this ACT-R extension successfully demonstrates the complex dynamics that emerge from interactions between physiology and cognitive components, it does not explain how those relations change over time.

Based on the above research, Nishikawa et al. \cite{nishikawa2022acs} constructed a model of a perceptual-motor coordination process accompanied by mind-wandering. By extending the previous studies \cite{Dancy:cn-BYwiN,van2015modeling}, their model defined arousal as the fluctuation (noise) of activation for the goals of the models. When the model falls into a low arousal state, the activation of the goal more fluctuated, and memory contents outside of the task tend to be more recalled. Following this, Nagashima et al. \cite{nagashima2022iccm} attempted to model the optimal level of arousal. This model leveraged the ACT-R module called {\it tracker}, which represents the nonlinear function of skill mastery \cite{fitts1964perceptual} and allows to simulate the process of acquiring motor skills in a rapid perceptual and motor coordination task \cite{anderson2019learning,Gianferrara2021CognitiveM}.

In summary, ACT-R has been utilized to create various cognitive function models in diverse contexts. However, prior research has primarily focused on the deficiency in task motivation/performance during low arousal situations. Notably, ACT-R has yet to be employed to model deficits in high arousal states. This study specifically targets the deficit induced by high arousal in the context of multitasking. 

\subsection{ACT-R Models Regarding Multitasking}
Several multitasking models employing ACT-R exist, elucidating a performance decline in such situations (see \cite{salvucci2008threaded}). However, most of these models do not integrate arousal dynamics. Rather, they explain the performance deficit in multitasking as a scheduling problem. For example, Kujala and Salvucci \cite{kujala2015modeling} constructed a model of dual-tasking, where drivers are required to manipulate the car and in-car displays simultaneously. In their model, the process associated with the subtask (manipulating the in-car displays) is executed only when the process associated with the main task (manipulating the car) is not executed. 
Although their model can explain the influence of the difficulty of the main task on the delay of the subtask process, it cannot simulate changes in the performance over time.

Another limitation of the above studies is that they only deal with situations where main and subtasks are always presented explicitly. Contrary to those concurrent multitasking, there is a situation where workers are required to deal with infrequent events that are independent of the main goal of the task. Such broad-sense multitasking was modeled in the aforementioned mind-wandering study \cite{van2015modeling}. In their research, a subgoal of probe response was posed to the model engaging a  sustained attention task. Nonetheless, their study does not detail performance changes in responding the subgoal stimuli. Thus, the present study attempts to describe the decline in overall task performance in this type of multitasking situation by building on the findings of the ACT-R studies presented so far.

\section{Task and Data}

\subsection{Line-Following Task}

\begin{figure}[tb]
    \centering
    \includegraphics[width=84mm]{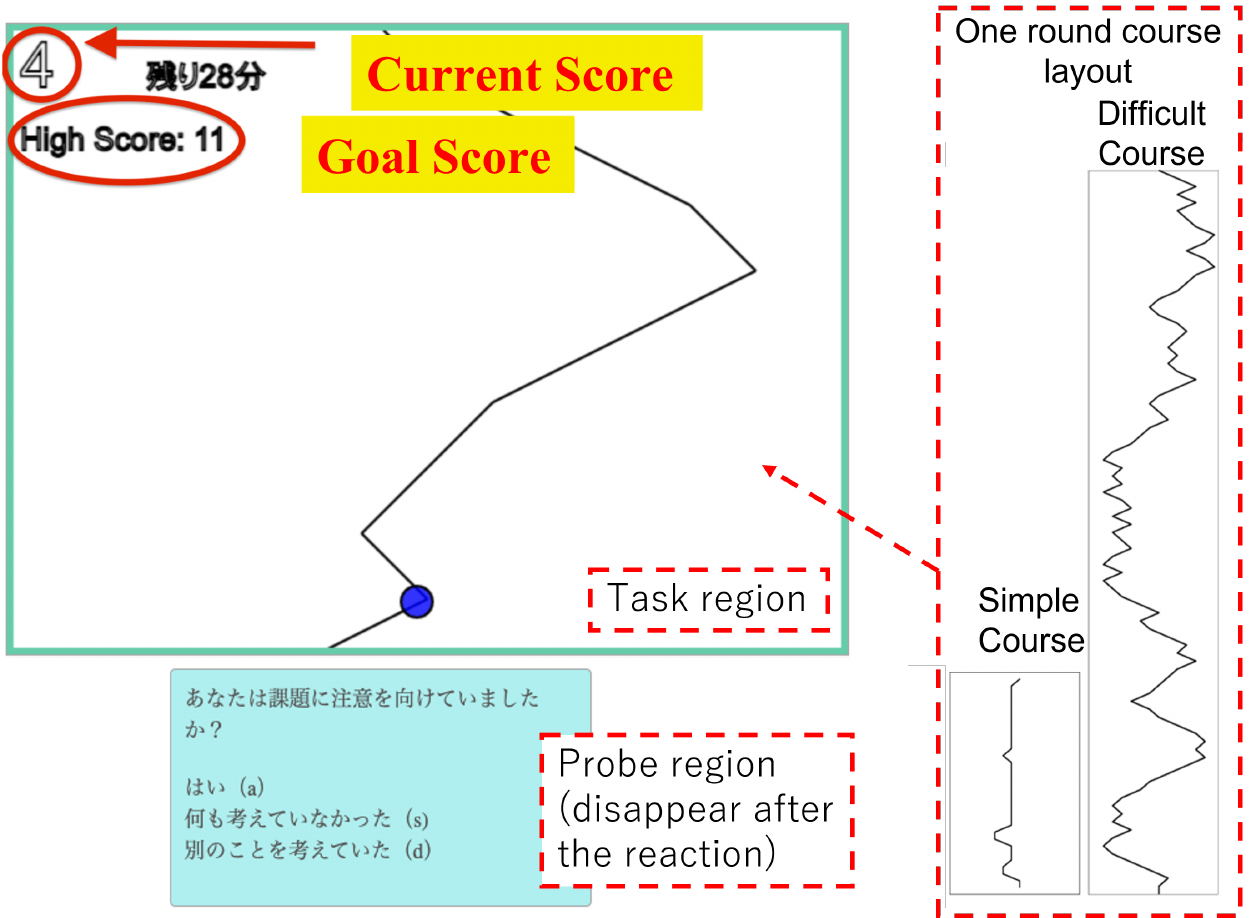}
 \caption{Task interface.} \label{fig:lft}
\end{figure}

We set up a line-following task \cite{maehigashi2013experimental}, which was used by Nishikawa et al. \cite{nishikawa2022acs}, as multitasking to examine the effects of arousal dynamics. Fig. \ref{fig:lft} shows the task interface. As the main goal of this task, participants operate the blue circle to follow a polyline that is displayed on the screen, automatically scrolling from top to bottom. The line varied according to a predefined course, as shown in the right panel of Fig. \ref{fig:lft}. These courses involve connecting patterns of lines with a height of 48 pixels at angles of 30, 45, 90, 135, and 150 degrees (0 degrees as the horizontal line).

The circle is moved by pressing the specified key. When a key is pressed during screen refresh, the circle moves two pixels in the direction indicated by the key. This specific amount of movement maintains the positional relationship between the line and the circle during the scrolling of the 30 and 150-degree lines. In other words, when the 30 and 150-degree lines continue, the key can be pressed continuously to successfully follow the lines. However, for a 45 or 135-degree course, intermittent key presses and releases are necessary to prevent excessive movement of the circle.

As a subgoal of this task, the participants are required to respond to probes (the bottom of Fig. \ref{fig:lft}) asking the state of concentration on the task. The probes are presented in average 50-sec intervals with noise which is randomly sampled from a normal distribution with a 5-sec standard deviation. Therefore, the number of times the probes are presented for the task is not constant. Also, if a participant does not respond before the next presentation cycle of a probe, that probe is skipped, and the next one is presented without interruption.

\begin{table}[tb]
\centering
\caption{Settings for the human experiment}\label{table:task_condition}
\scalebox{0.92}[0.92]{
\begin{tabular}{c|c|c|c}
\hline
& \begin{tabular}[c]{@{}c@{}}Course\\ (frame)\end{tabular}     & \begin{tabular}[c]{@{}c@{}}Scroll speed\\ \end{tabular} & \begin{tabular}[c]{@{}c@{}}Goal\\ setting\end{tabular} \\ \hline
\begin{tabular}[c]{@{}c@{}}Low arousal\\ demand condition\end{tabular} & \begin{tabular}[c]{@{}c@{}} Simple\\ (1500 frame)\end{tabular}      & 40 ms/frame                                                        & Nothing   \\ \hline
\begin{tabular}[c]{@{}c@{}}High arousal\\ demand condition\end{tabular}  & \begin{tabular}[c]{@{}c@{}} Difficult\\ (4500 frame)\end{tabular} & 13 ms/frame        & Score  \\ \hline
\end{tabular}
}
\end{table}

%In the task, we designed two line-following exercises with different difficulty levels: one aimed at inducing a low arousal level (low arousal induction condition) and the other at inducing a high arousal level (high arousal induction condition). For the low arousal condition, we set monotonous tasks with low difficulty for typical participants, while the high arousal condition involved challenging tasks that could only be completed with higher arousal levels. 
In the experiment, the difficulty of the task was adjusted by manipulating parameters such as the ratio of vertical lines in the line pattern and the scrolling speed. Specifically, the former parameter was represented by the simple and difficult courses in the right panel of Fig. 1. For both courses, the scrolling speed was set to complete one round in one minute. The resulted parameters are shown in the second and third columns of Table \ref{table:task_condition}.  We assume that these different conditions require varying levels of arousal. Therefore, in this paper, we refer to the condition where a simple course scrolls at a slow speed as the ``low arousal demand (LAD) condition'' and the other as the ``high arousal demand (HAD) condition.'' These labels indicate that the conditions do not necessarily reflect participants' actual arousal levels but instead represent the arousal levels required for the task.

In addition to the task difficulty, we manipulated goal setting. Based on a theory of motivation \cite{alter2017irresistible}, the arousal level for a task is assumed to be influenced by this factor. Consequently, in the HAD condition, participants were presented with the highest score achieved in the previous rounds up to that point in the task, and they were instructed to aim to surpass that score. The score here indicated the percentage of frames in which the line was successfully traced during the round. 

\subsection{Human Behavioral Data}\label{human_data}
% One hundred participants were recruited through a Japanese crowdsourcing site (Lancers.jp) and divided into two groups corresponding to high or low arousal induction conditions. 

The data were collected through a Japanese crowdsourcing site (Lancers.jp).
We posted a call for up to 50 participants on the crowdsourcing site for each condition (100 participants, including both conditions). 

In both conditions, the participants were forwarded from the crowdsourcing site to a website on the authors' server. At the site, they engaged in the line-following task for 30 min after entering their user identification of the crowdsourcing service and reading the instructions for the task on the screen. At the instruction page, they were posed quizzes to confirm their understanding of the instructions. 

The call was closed when the number of participants registered on the crowdsourcing site reached the limit. Then, we used all the data on the server, which recorded valid user identification and 30-min behavior logs of the task (position of the circle and lines for each frame, timing of probe presentations, and responses).
As a result, we obtained the data from 24 participants in the LAD condition and 39 in the HAD condition for the target of modeling.

%We analyzed data that included 30 minutes of logs and responses to the questionnaire. The data were obtained from 24 participants in the low arousal induction condition and 39 participants in the high arousal induction condition.
% Excluding participants with incomplete data, we analyzed data from 24 participants in the low arousal induction condition and 39 participants in the high arousal induction condition. 

\begin{table*}[tb]
\centering
\caption{Regression results of human data} \label{table:human_regression}
\begin{tabular}{c|c|c|c|c|c|c}
\hline
Indicator & Condition & Variable & Coefficient & SE & t & P \\ \hline
\multirow{6}{9em}{Offline Ratio} & \multirow{3}{1.5em}{LAD} & Intercept & 0.0487 & 0.007 & 7.389 & $<$.0001**\\
& & $x$ & -0.0052 & 0.001 & -4.970 & $<$.0001**\\
& & $x^2$ & 0.0001 & $<$.0001 & 3.995 & $<$.0001**\\ \cline{2-7}

& \multirow{3}{1.5em}{HAD} & Intercept & 0.3328 & 0.009 & 35.778 & $<$.0001**\\
& & $x$ & -0.0085 & 0.001 & -5.737 & $<$.0001**\\
& & $x^2$ & 0.0002 & $<$.0001 & 4.156 & $<$.0001** \\ \hline

\multirow{6}{9em}{Response Time} & \multirow{3}{1.5em}{LAD} & Intercept & 3.1077 & 0.275 & 11.318 & $<$.0001** \\
& & $x$ & -0.1366 & 0.044 & -3.117 & 0.004** \\
& & $x^2$ & 0.0036 & 0.001 & 2.439 & 0.022* \\ \cline{2-7}

& \multirow{3}{1.5em}{HAD} & Intercept & 2.7691 & 0.362 & 7.651 & $<$.0001** \\
& & $x$ & 0.0588 & 0.058 & 1.018 & 0.318 \\
& & $x^2$ & -0.0018 & 0.002 & -0.939 & 0.356 \\ \hline

\multirow{6}{9em}{Response Time (STD)} & \multirow{3}{1.5em}{LAD} & Intercept & 3.5977 & 0.555 & 6.482 & $<$.0001** \\
& & $x$ & -0.2359 & 0.089 & -2.663 & 0.013*\\
& & $x^2$ & 0.0067 & 0.003 & 2.270 & 0.031*\\ \cline{2-7}

& \multirow{3}{1.5em}{HAD} & Intercept & 3.6245 & 1.142 & 3.173 & 0.004** \\
& & $x$ & 0.2876 & 0.182 & 1.577 & 0.126 \\
& & $x^2$ & -0.0076 & 0.006 & -1.253 & 0.221 \\ \hline
\end{tabular}
\begin{tabular}{ccccccc}
\multicolumn{7}{l}{Note. LAD represents the low arousal demand condition in humans. }\\
\multicolumn{7}{l}{ HAD represents the high arousal demand condition. }
\\
\multicolumn{7}{l}{ *: $p<.05$, **: $p<.01$. }
\end{tabular}
\end{table*}

\begin{figure}[tb] 
    \centering
    \includegraphics[width=84mm]{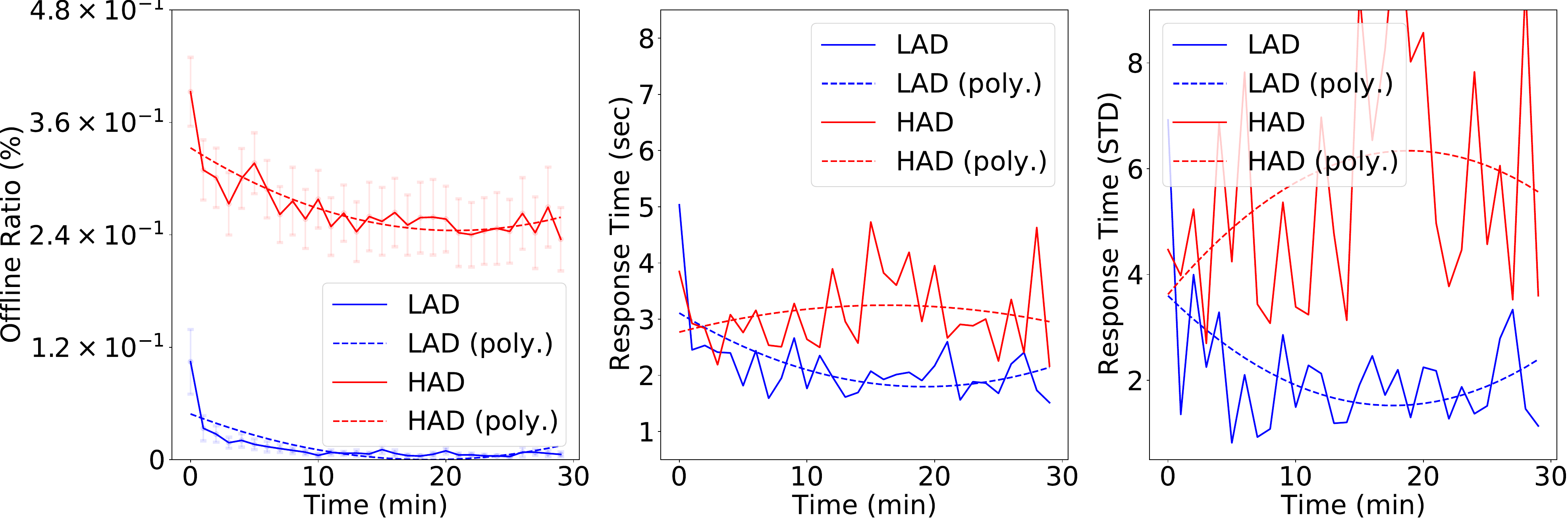}
    \caption{Human data. The x-axis represents 30 time points (one-min rounds). The y-axis represents the time-series change in each indicator averaged over participants  ($n=24$ for LAD, $n=39$ for HAD). Each dotted line indicates the result of a polynomial regression (degree = 2) shown in Table \ref{table:human_regression}. Error bars attached to the offline ratio indicate a standard error of mean.} \label{fig:human}
\end{figure}

The trends of the performance of the main and sub goals for the LAD ($n=24$) and HAD ($n=39$) conditions are summarized in Fig. \ref{fig:human}. The horizontal axes represent 30 time points corresponding one-min rounds, while the vertical axes indicate the performance of the main and sub goals, averaged across participants in each condition. %For the main goal, we averraged  and approximately 50 seconds intervals for the sub goal) of a task run. 
The performance of the main goal is represented as logarithmic offline ratios (the percentage of frames where the circle did not follow the line out of the total frames in a round). In contrast, subgoal performance is measured by probe response time (the time between the probe's appearance and the button press) for each instance of the probe. For the probe response time, we separately presented the means and the standard deviations across the participants because we considered that overlooking subgoal-related stimuli is reflected in the variance of the response time. 

As a preprocessing step, we calculated the average of the probe responses when multiple responses were recorded within the same round. Additionally, if participants did not respond to a probe before the next one appeared, the response time for that probe was recorded as 50 sec, representing the expected maximum value. The subsequent probe following the non-responded probe, which appeared continuously on the screen, was treated as missing data and was not included in the statistical analysis. While there were no missing data in the LAD condition, 12 probe responses were treated as missing in the HAD condition.

For each measure of the performance in figures, we conducted a quadratic regression analysis by setting a significance level at 0.05 ($n=30$ using the data points in the figure as units of analysis). The obtained coefficients for each condition are summarized in Table \ref{table:human_regression}. Fig. \ref{fig:human} also shows the trends calculated from the obtained coefficients. From the analyses, we can confirm the significant coefficient of the time ($x$, $x^2$) in the offline ratio of both the LAD and HAD conditions, indicating the improvement of the performance of the main goal over time. Contrary, concerning the subgoal performance (response time and its STD), we find the significant coefficient of the time in only the LAD condition.  The results suggest that subgoal performance improves only under the LAD condition, not under the HAD condition. 

We also examined the differences in the performance of the main and sub goals between conditions. In Table 2, coefficients of the intercepts for offline ratio are clearly different between the LAD and HAD conditions (0.0487 $<$ 0.3328, respectively). From this difference, we can confirm the influence of the task difficulty on the main goal performance. 
In contrast, the difference in performance on the subgoal is not clear in the table, although from the middle and right panels of  Fig. 2 we can observe the  difference between the two conditions in the means and STDs of response time. To confirm these partial differences, we performed a paired t-test using the values at the data points in the figure as the unit of analysis ($n=30$). For the middle of Fig. 2, we confirmed the difference between the average of the response time between the conditions ($t (29) = 5.94$, $p<.01$). We also confirmed similar differences in STDs presented in the right panel of Fig. 2. This suggests that the probe response in the HAD condition were generally delayed (the middle panel of Fig. 2). In particular, certain rounds by some participants in the HAD condition experienced significant delays in responding to the probe as shown in the right panel of Fig. 2.

To summarize, the characteristics of the time trends for each dataset are as follows:
\begin{itemize}
    \item Offline ratio (main goal) in LAD: decreasing over time
    \item Offline ratio (main goal) in HAD: decreasing over time
    \item Response time (subgoal) in LAD: decreasing over time
    \item Response time (subgoal) in HAD: no statistically significant time trends
\end{itemize}
Additionally, we observed the following differences between conditions:
\begin{itemize}
    \item Offline ratio (main goal): HAD $>$ LAD
    \item Response time (subgoal): HAD $>$ LAD
\end{itemize}

Our model aims to reproduce these characteristics of human data, particularly focusing on the probe response time as the subgoal indicator.

\section{Model}
The current study extended the previous model of arousal in the line following task \cite{nagashima2022iccm,nishikawa2022acs} to perform the task in a multitasking situation. As noted in Section 1, the model was implemented in ACT-R \cite{Anderson:2007}, which is the representative cognitive architecture. One of the important assumptions of ACT-R is that the human cognitive process has two aspects: symbolic and subsymbolic ones. The former can be represented as discrete states, while the latter modulates the former states based on numerical parameters. Like the other studies of ACT-R, our model can also be explained along with these aspects as follows. 

\subsection{Symbolic Process} \label{sec:symb}
\begin{figure}[tb]
\includegraphics[width=84mm]{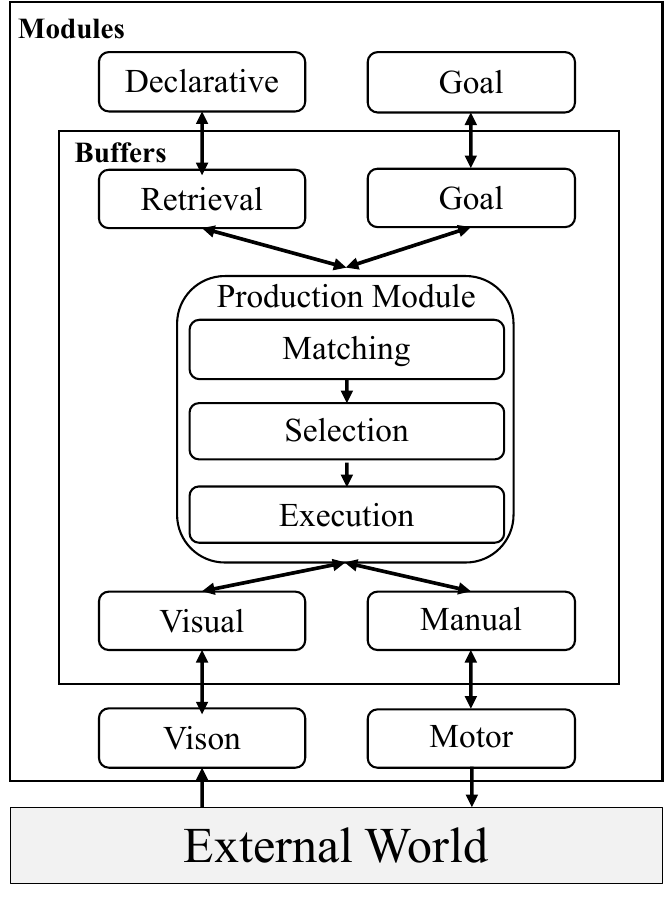}
\centering
\caption{
Modules of the adaptive control of thought-rational (ACT-R) used in the model. This figure is created with reference to \cite{anderson2004integrated} and \cite{Ritter:inpress}.
}.

\label{modules}
\end{figure}

ACT-R represents the states in each moment of task execution as a set of {\it chunks}\footnote{Chunks in ACT-R is a data structure consisting of multiple slots to store each piece of data.}, which is a symbolically encoded information used in the task. During the task execution, each chunk is stored in a buffer associated with a module, which corresponds to a brain region playing a functional role in the task. Among several modules in ACT-R, the vision, motor, goal, declarative, and production modules are used for the line-following task. Those modules are structured as shown in Fig. \ref{modules} and function as follows.

\subsubsection{Vision Module}
This module simulates interaction with the external environment by recognizing information needed to perform a task. To instantiate the interaction, ACT-R provides a system called ``ACT-R graphical user interface (AGI)'', which places information required to conduct a task in a two-dimensional space. 

For the line following task, we created a virtual window of AGI, placing symbols representing the circle and lines. From this window, the model obtains a positional relationship of the circle to the line to be followed. In addition, the model confirms the appearance of the probe through the module.
The obtained information by the vision module, namely a chunk, is placed on the buffer associated with the module and transferred to the other modules.

\subsubsection{Goal Module}
This module stores chunks representing the current goal and the current state of the task to achieve the goal. In this study,  the module stores the chunk representing the current goal (the main and sub goals) and the current state to achieve those goals in the line-following task. The goal module also stores tentative information obtained from the vision module, such as the circle position and next turn position, and a {\it flag} indicating whether the model has seen the probe or not. 

\subsubsection{Motor Module} \label{motor_module}
This module simulates the physical operations required by the task. In the line-following model, key presses are performed to move the vehicle. Nagashima et al. \cite{nagashima2022iccm} prepared four operations:  {\it Stop} (release the key), {\it Left} (press the key assigned to the left), {\it Right} (press the key assigned to the right), and {\it Continue} (continue the previous operation). In addition to those operations, this study adds  {\it Left Punch} (briefly press the key assigned to the left twice) and {\it Right Punch} (briefly press the key assigned to the right twice) to follow 45 and 135 degrees angle lines.

\subsubsection{Declarative Module}
This module simulates declarative memory, such as episodic memories and semantic knowledge. In other words, this module plays a database permanently stores information used for the model. In the context of multitasking, it should be emphasized that the task goals are also stored in this module. Thus, in the task, the model is required to retrieve the chunk representing the current goal (main and sub goals) from the declarative module while monitoring the current external environment. The retrieved goal is stored in the buffer associated with the module and then transferred to the goal module by leveraging the production module, as noted below.

\subsubsection{Production Module}
\begin{figure}[tb]
    \centering
    \includegraphics[width=84mm]
    {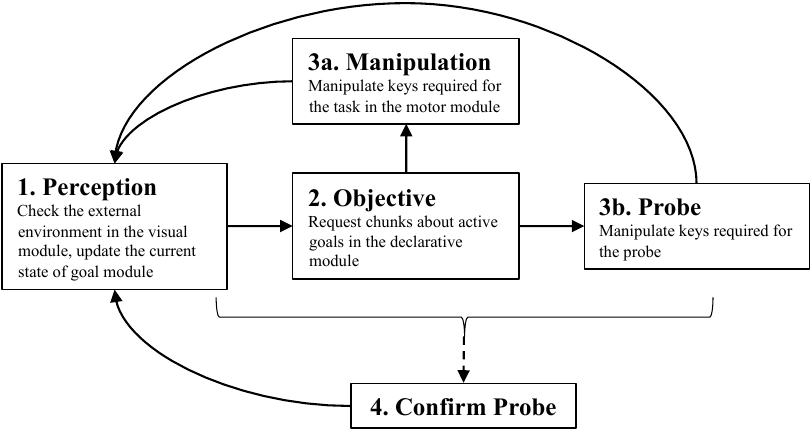}
 \caption{Block diagram showing basic model processing.} \label{fig:state}
\end{figure}

Communication between the above modules is mediated by production rules stored in production modules. In each step of the model, one production rule is selected by refereeing to the states of the model (chunks in the buffers). The selected production rule takes action such as transferring information between buffers (e.g., copying information from the visual buffer to the goal buffer) and sending a command to the modules (e.g., changing the operation of the motor module to ``Left''). Connecting such individual production rules makes a process to execute a task. In this study, following the previous model of arousal in the line-following task \cite{nishikawa2022acs,nagashima2022iccm}, we constructed the cyclic process as shown in Fig. \ref{fig:state}. Each box in the figure corresponds to a production rule described as follows\footnote{The difference from the previous study is that the probe confirmation (\textit{Confirm Probe} state) and the probe reaction (\textit{Probe} state) were separated. In previous studies, probe confirmation and manipulation were performed simultaneously. In the current study, we constructed the multitasking situation by separating this behavior.}. 

\begin{enumerate}
    \item \textit{Perception}: This production rule transfers the positional information of the circle and the lines stored in the visual buffer to the goal buffer.
    \item \textit{Objective}: This production rule sends a command to retrieve a chunk representing the goals (main or sub) to the declarative module. The selection of retrieving chunks from the declarative module is governed by the subsymbolic process explained in the next subsection.
    \item \textit{Motor actions}: Based on the current goal, the model takes one of the following production rules.
\begin{enumerate}
        \item \textit{Manipulation}: When the goal buffer stores the main goal, the model chooses one of the production rules responsible for operations noted in \ref{motor_module} to trace the line with the blue circle. The selection of the rules depends on the positional information stored in the goal buffer (e.g., if the circle is on the left of the line, then ``Right'' is selected). We also utilize the tracker module, which is responsible for motor learning \cite{anderson2019learning,Gianferrara2021CognitiveM}, to adjust the boundaries of selecting these actions (e.g., how many pixels between the circle and the line are required to judge ``on the right''). The details of the mechanism is described in Appendix.
        \item \textit{Probe}: When the goal module stores the sub goal, the model selects a production rule to send a command pressing a space bar to the motor module to react a probe. The rule also changes the flag in the goal buffer, indicating whether the model has seen the probe to ``off'' after pressing the space bar.%In this state, the model is transitioned to \textit{Confirm Probe} state described below.
\end{enumerate}
    \item \textit{Confirm Probe}: This production rule is not directly connected with the above process. Instead, it is triggered by a chunk showing the probe in the visual buffer. Storing the probe chunk in the visual buffer is automatically executed by a function of ACT-R called ``buffer staffing.'' Once the rule is selected, the flag in the goal buffer indicating the model has seen the probe is changed to ``on.'' This flag information indirectly affects the retrieval of the next goal (``objective'' rule) through the subsymboilc process explained in the next section.
\end{enumerate}

After the above steps, the vision module checks for a new state in the external environment and returns to the first step. The time required for the above process is determined by the method defined in ACT-R (e.g., production execution time of 50 ms). 

\subsection{Subsymbolic Process} \label{sec:subs}
The above process varies depending on the recalled chunk representing the task goal. In ACT-R, such retrieval process is modulated by a parameter called ``activation.'' ACT-R calculates activation for each chunk when the retrieval request is made. The chunk with the highest activation is recalled for the request. This sub section describes the ACT-R's computation of activation with our original modifications to simulate high arousal states.

\subsubsection{Activation and Probe Response}
\begin{figure}[tb]
    \centering
    \includegraphics[width=84mm]{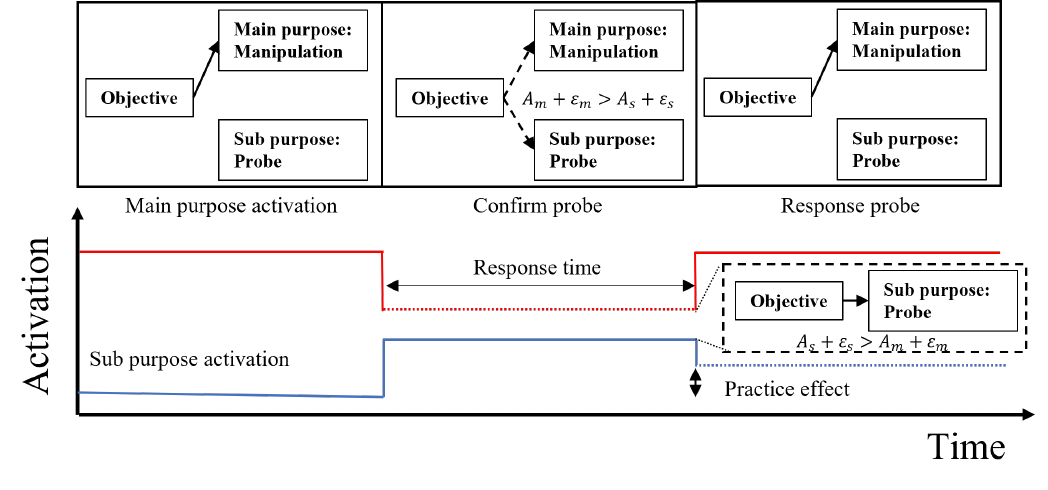}
 \caption{Schematic explanation on the changes in activation in the probe response. The red line represents the activation of main goal and the blue line represents the activation of subgoal. The solid line represents the activation with the effect of spreading activation added, and the dotted line represents the activation without the effect.} \label{fig:activation}
\end{figure}

According to the explanation so far, the strength of concentration on the task can be modeled as the activation of chunks representing goals. Based on this basic idea, Fig. \ref{fig:activation} illustrates task switching based on the activation dynamics for the main and the sub goals. Generally, the activation for the main goal is assumed to be higher than that of the subgoal because of the emphasis on the task instruction. However, when a stimulus related to the subgoal is presented, it is expected that the activation of the subgoal will temporarily increase and close to the activation of the main goal. When the activation of the subgoal exceeds the activation of the main goal, the chunk stored in the goal buffer is changed, and a response to the presented stimulus is made. Then, after the stimulus disappears, the activation of the subgoal decreases and returns to the main goal.

The above process can be represented by leveraging the basics of ACT-R subsymbolic computations. First, the activation ($A_i$) of chunk $i$ in ACT-R is calculated as the summation of multiple components.
\begin{equation}
\label{activation_f}
A_i = B_i+S_i+\varepsilon_i
\end{equation}

The above equation only contains components used in the current study. $B_i$ is called base-level activation, representing learning and forgetting effects. Specifically, $B_i$ in the current study is calculated by
\begin{equation}
\label{baselevel_f}
B_i = \ln (num /(1-d))-d*\ln (L)+\beta
\end{equation}
where $num$ is the number of times each chunk is presented, $L$ is the elapsed time since the first appearance of the chunk, and $d$ is the decay rate and $\beta$ is an offset value applied to all chunks in the model\footnote{This equation is an approximation of the original equation of ACT-R.}. Thus, this component is changed by how many times the goal is repeated in the instruction and how frequent the goal is activated in the task.

The second term of equation \ref{activation_f}, $S_i$, is called spreading activation and is used to represent the contextual effect on the activation. In ACT-R, the context indicates the buffer-contents at the moment of retrieving the chunk. Thus, spreading activation indicates an implicit effect of chunks in the model on the retrieval process.
More specifically, $S_i$ is calculated by 
\begin{equation}
\label{spreading_f}
S_i = \sum_{j}W_{j}S_{j, i}
\end{equation}
where $j$ denotes the distinction between slots in the buffer. $W_j$ is the weight of the activation, which is usually a value divided one by the number of $j$. $S_{j, i}$ is the parameter determining the strength of the spreading activation from buffer-content $j$ to chunk $i$. As indicated in Section \ref{introductin} and Fig. \ref{modules}, each module has a dedicated buffer. Through this, the modules send and receive chunks to and from the production module. The current study only targets the contents of the goal buffer and makes an association with the flag of observing a probe to the chunk representing a subgoal.

The last term of equation 1, $\varepsilon_i$, is a tentative noise, which is stochastically determined by a logistics distribution based on 0 and is considered a component that varies independently of the other activation components. Thus, when the noise is large relative to the other components of activation, transitions between goals are frequent, and when the noise component is small relative to the other activation value, sustained activity for a single objective is made. 

The switching between goals in this study was especially designed by utilizing the spreading activation and noise.  After the probe confirmation, as shown in Fig. \ref{fig:activation}, the $S_{Subgoal}$ is increased because of the additive effect from the newly added relevant buffer contents ($S_{flag, Subgoal}$ = 1) while \textbf{$S_{MainGoal}$} is decreased because of the proportionally decreasing effects from the irrelevant new buffer content ($S_{flag, MainGoal}$ = 0). 

However, we assume that only the spreading activation is not enough to switch from the dominant main goal to the rarely occurred sub goal. As shown in Fig. {\ref{fig:activation}}, the main goal obtains more base-level activation than that of the subgoal, and a gap exist between the subgoal relative to the main goal even after the probe confirmation. We then considered that the gap here is filled by the noise temporarily assigned to the activation. We supposed that the noise value assigned to the two chunks causes the activation of the subgoal to temporarily exceed the activation of the main goal, and the effect of the noise is reduced in high arousal situations as discussed in the following section.

\subsubsection{Activation in High Arousal State}\label{reward}
In this study, we regard the performance decline in high arousal as a lack of smooth transition of activity from the main goal to the subgoal. To represent the arousal dynamics that interfere with the allocation of attention to the subgoal, we redefined the activation equation by modifying equation 1 as follows.

\begin{equation}
\label{activation_2f}
A_i = r(t)(B_i+S_i)+\varepsilon_i
\end{equation}

\begin{figure}[tb]
    \centering
    \includegraphics[width=84mm]{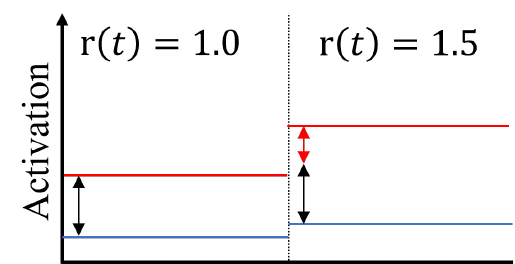}
    \caption{Change in activation for each chunk in the coefficient application. The red line represents the main goal activation and the blue line represents the subgoal activation. The black arrow represents the difference in activation at $r(i)=1.0$, and the red arrow represents the incremental difference in both activation raised by the high arousal state.}\label{fig:reward}
\end{figure}

This new equation introduces a coefficient, $r(t)$, that is applied to the activation of all chunks of the model (including both chunks representing the main and the sub goals). Fig. \ref{fig:reward} illustrates the changes of the activation for the main and sub goal chunks in the different $r(t)$ values.
The basic idea behind this is that a highly aroused situation makes memory access rapid and easy. The ACT-R's activation mechanism not only affects the selection of a chunk but also influences the access speed of the chunk. Highly activated chunks are retrieved fast, and if there are no chunks with enough activation, the model fails the memory retrieval after the predefined threshold time passed. Thus, we assume that in extremely low arousal situations, the total process of the model is delayed\footnote{A similar assumption was made by Juvina et al. \cite{JUVINA20184}, who proposed a core affect model in ACT-R. In their model, arousal is assumed as a by-product of the valuation process and affect total activation values. The difference is that to affects the total activation, including $S_i$, our model represents the effect as a coefficient rather than an independent term.}.

In our study, such a coefficient affecting total activation is decided by the reward based on the degree of the task successes. In the current study, we specifically define $r (t)$ as
\begin{equation}
\label{reward_f}
r(t)= \alpha(100 - | score_{goal} - score_t | ) / 100
\end{equation}
where $score_{goal}$ is the target online ratio, $score_i$ is the current online ratio, and $\alpha$ is the scaling constant. That is, $r(t)$ increases as the current score approaches the goal score and decreases as the score departs from the goal score. We defined this equation based on predictive coding \cite{clark2013whatever,friston2010free,rao1999predictive,SETH2013565}. This reward is higher the smaller the difference between the target score ($score_{goal}$) and the current score ($score_t$) is. In other words, the model is biased towards minimizing the difference between its own prediction and the current state. 

Note that in the HAD condition in this study, participants are required to keep updating their scores in 30 rounds. That is, $score_{goal}$ is the best score in the previous rounds. In addition, the model includes motor learning components \cite{nagashima2022iccm}. In other words, the goal score is updated as motor learning progresses through the task. This means that the model's goal in the task will always require a higher level of motor learning, and the model maintains a state of high arousal throughout the task.

\begin{figure}[tb]
    \centering
    \includegraphics[width=84mm]{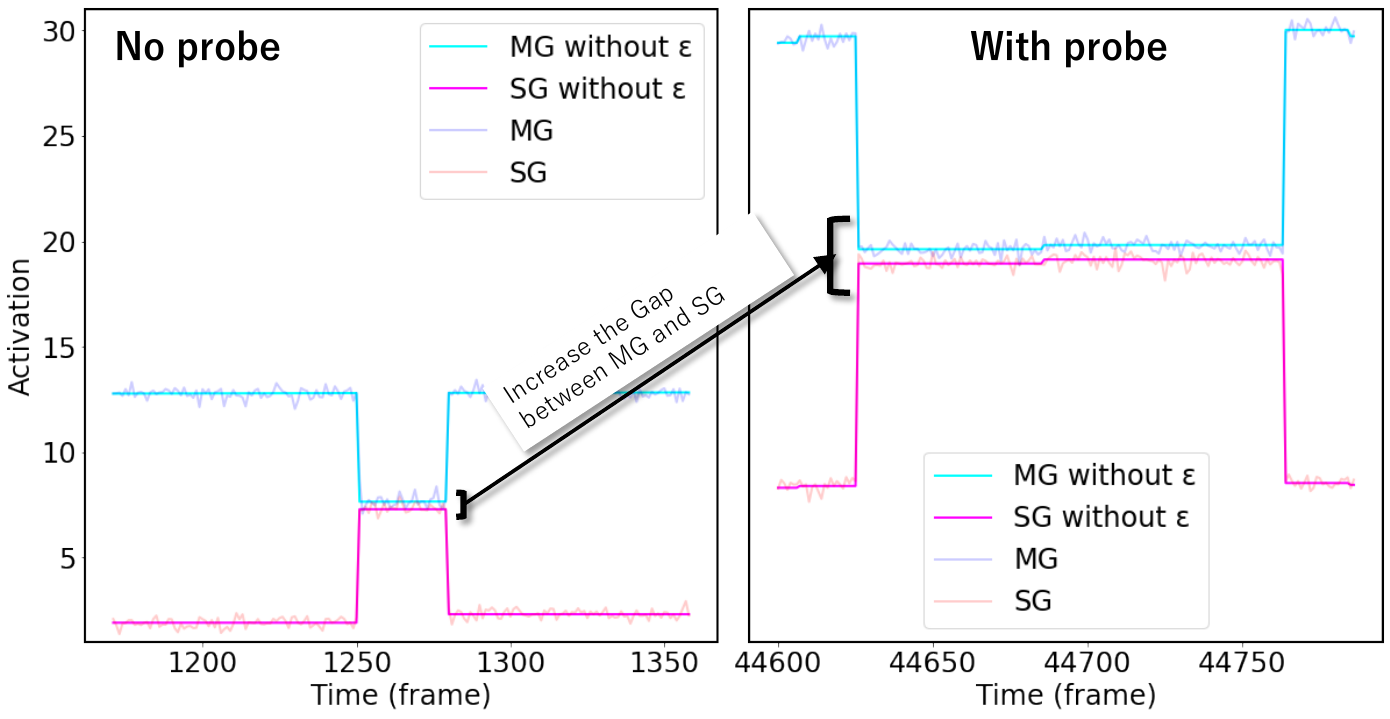}
    \caption{Examples of activation values. The x- and y- axes indicate the frame corresponding to the time (1 frame per 40ms) and the activation respectively. MG: main goal, SG: subgoal.} \label{fig:ra}
\end{figure}

In addition, since the noise term is independent in equation 4,  the effect of noise is relatively smaller at high arousal than at low arousal situations. Fig. \ref{fig:ra} shows an actual example of the change in the activation of the main goal and subgoal obtained in a same run of the model. The left and right graphs show the situation when $r(t)$ value is small and large, respectively. Since $r(t)$ is defined as a coefficient on the activation, the absolute difference between the activation of the main goal and subgoal (the gap between the blue and the pink lines) in the low arousal state is smaller than that in the high arousal state\footnote{In contrast to the left graph, the right graph subtly decreases the difference between the activation values of the main goal and the subgoals after the probe presentation.
Although this difference seems small, the graph on the right significantly delays the reaction time to the probe due to the relationship with the fixed noise size.}. This relationship explains why, compared to in the low arousal state, the activation of the subgoal in the high arousal state is less likely to temporarily exceed the activation of the main goal; due to the relatively small amount of noise fluctuations in the high arousal state, reaction time delays.

\section{Simulation}
\subsection{Aim and Settings}
The simulation in this study examines the behavior of the above models in settings that correspond to the LAD and HAD conditions in the human experiment. The model traces the course that participants followed in each condition for 30 min at the same speed as in the LAD condition (40 ms/frame). We also set two models manipulating the goal setting in equation \ref{reward_f} as follows.

\begin{itemize} \label{common_setting}
    \item{Low arousal demand model (mLAD): As with the LAD condition in the experiment, no feedback by score was given and the arousal level of the model was assumed to be fixed to a neutral state. That is, $r(t)$ was fixed to 1 throughout the task. The activation of the main goal and the subgoal remained computed by the default ACT-R setting.}
    \item{High arousal demand model (mHAD): The goal score is updated as in the human experiment. Since the model is in a situation where the goal score is constantly tracked, $score_{goal}$ fluctuates with the performance from round 2 onward.  $\alpha$ is set so that the average of $r(t)$ calculated by this fluctuation is greater than 1. In the initial round, however, $r(t)$ is set to 1 because there is no goal score. }
\end{itemize}

In addition to these settings, we set $\sigma = 0.13$, $\beta_i = 4$, $d = 0.5$, and $S = 16.84$ as the ACT-R parameters for memory activation. The parameters of the motor module were set the same as those of the previous study \cite{Gianferrara2021CognitiveM}, and the parameters of motor learning were also set the same as those of the other previous study \cite{nagashima2022iccm}.

\begin{table}[tb]
\centering
\caption{Set of parameters for each chunk.} \label{table:parameter}
\begin{tabular}{c|cc|cc}
\hline
                       & \multicolumn{2}{c|}{Main goal}    & \multicolumn{2}{c}{Subgoal} \\ \hline
\multicolumn{1}{l|}{} & \multicolumn{1}{c|}{num}    & L    & \multicolumn{1}{c|}{num} & L    \\ \hline
Parameter 1                 & \multicolumn{1}{c|}{1800} & 1800 & \multicolumn{1}{c|}{5} & 1800 \\ \hline
Parameter 2                 & \multicolumn{1}{c|}{500}  & 1800 & \multicolumn{1}{c|}{2} & 1000 \\ \hline
\end{tabular}
\end{table}

Table \ref{table:parameter} shows two sets of initial parameters related to the baseline activation ($num$ and $L$). Those settings were intended to reflect the instructions made before the task, which emphasized the main goal over the subgoal during the experiment. Consequently, the number of times ($num$) the subgoal chunks were presented (including how many times they were seen and recalled during repeated reading of the instructions) was intentionally kept lower than the chunks for the main goal. Additionally, assuming that these chunks were introduced for the first time at the instruction of the experiment, we set a baseline of $L$ as 30 min (1800 s). The two parameter sets were based on different assumptions. Parameter 2 featured a smaller number of presentations (repetitions) compared to parameter 1. Furthermore, the chunks of the subgoal were constructed as more recent memories (mentioned as the subgoal at the end of the instructions).

\subsection{Result} \label{result}
\begin{figure}[tb]
    \centering
    \includegraphics[width=84mm]{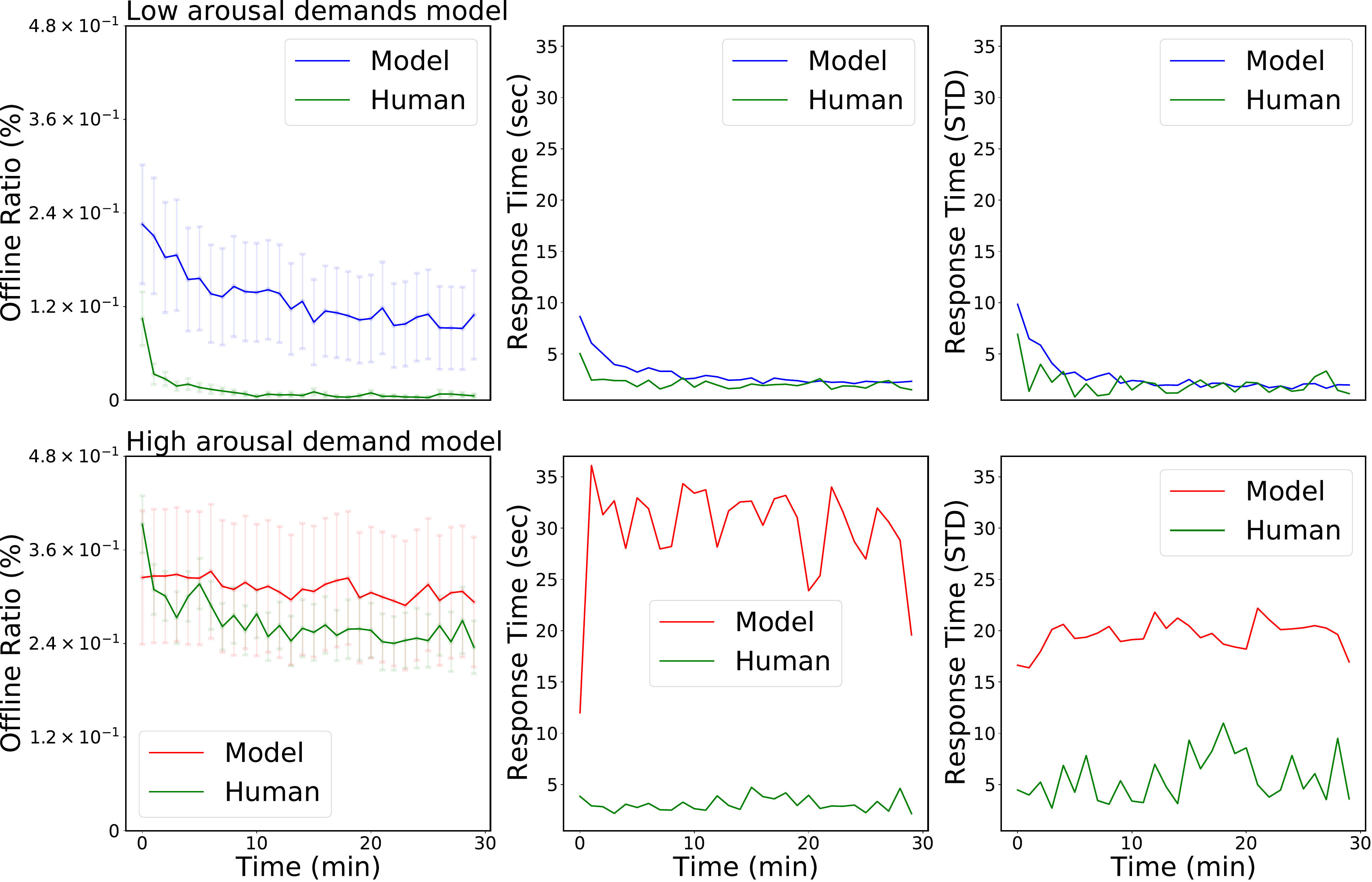}
    \caption{Time-series changes of each indicator in the simulation with parameter 1 ($n=150$). Model indicates the result of the simulation and Human referes the result of the human experiment shown in Fig. \ref{fig:human}. Error bars represent a standard error of means.} \label{fig:p1}
\end{figure}

\begin{figure}[tb]
    \centering
    \includegraphics[width=84mm]{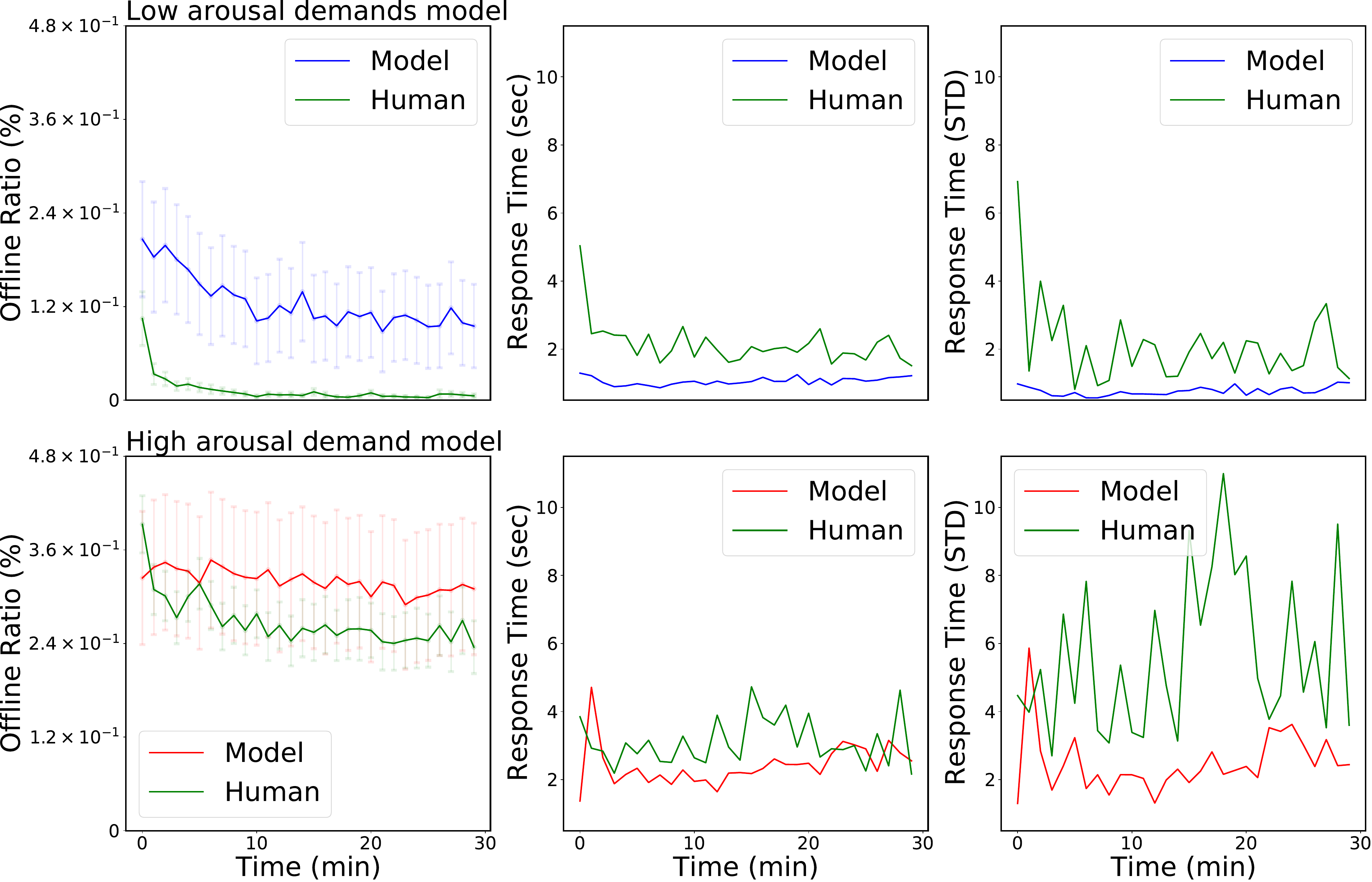}
 \caption{Time-series changes of each indicator in the simulation with parameter 2 ($n=150$).} \label{fig:p2}
\end{figure}

With the above parameter sets, the consistency of the behavior between the model and the experiment was investigated. 
Each model with each parameter was executed 150 runs to calculate averages of the offline ratio and the response time to the probe as shown in Figs. \ref{fig:p1} and \ref{fig:p2}. 

The left panels of those figures represent the offline ratios under all conditions and parameters. Despite the consistency in the general decreasing trends with the human data, the value in the low arousal model is significantly lower than that in the corresponding human data. Those results suggest the benefit of the ACT-R motor learning mechanism \cite{anderson2019learning} to simulate the improvement of the main goal performance, although the perceptual-motor cycle in ACT-R is indicated as slower than that in the human participants.

Contrary, the middle and right panels of Figs. \ref{fig:p1} and \ref{fig:p2} show the results relating the probe response time, which is the focus of this study. 
In generating these figures, we set the response time for unresponded probes to 50 sec, similar to the human data, and treated subsequent responses as missing data.
As a result, the models with parameter 1 generated no missing data, while the model with parameter 2 generated one missing response for the mLAD model and 3,149 missing responses for the mHAD model. Thus, the mHAD model with parameter 2 failed to respond to more than half of the probes, and the almost half of the remaining probes were assigned the maximum response time of 50 sec. This resulted in a large average with significant standard deviations as shown in Fig. 8. 
Thus, for parameter 1, we can observe that the model's reaction time is higher than the human data. In contrast, for parameter 2, the model's reaction time is lower than the human data.

\begin{figure}[tb]
    \centering
    \includegraphics[width=84mm]{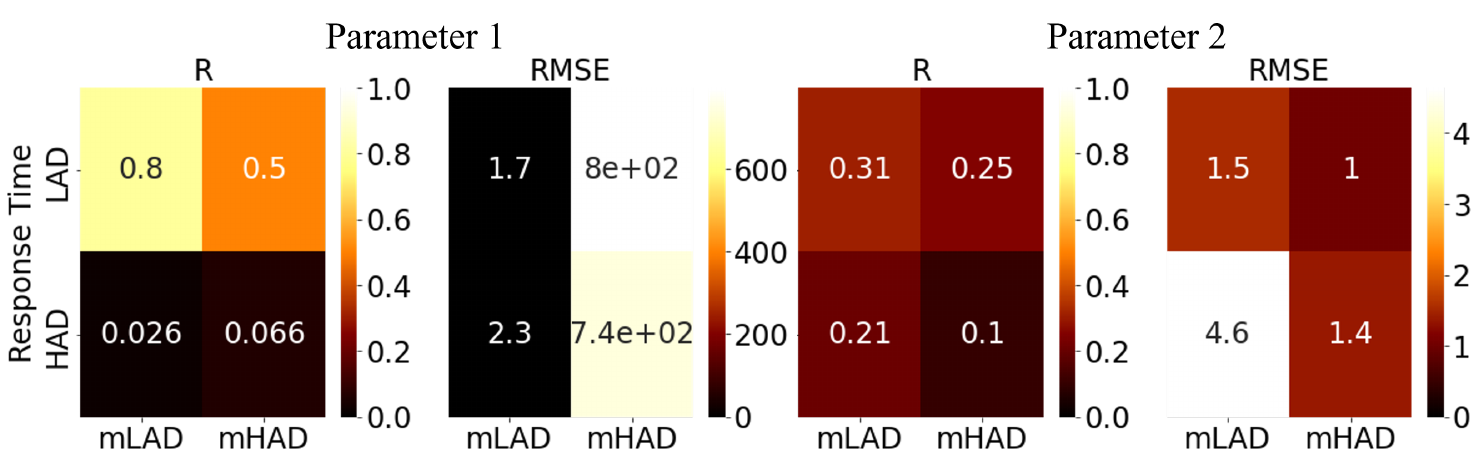}
    \caption{Model-human fit for the subgoal. mLAD represents the low arousal demand model, and mHAD represents the high arousal demand model.}
    \label{fig:p1p2_m}
\end{figure}

To examine details of the differences in these trends, the fitting of each model and parameter to human data was examined using the correlation coefficient (R) and RMSE (Root Mean Square Error) for the averages of the probe response time in the middle panels of Figs. \ref{fig:p1} (parameter 1) and \ref{fig:p2} (parameter 2) as units of the analysis. Fig. \ref{fig:p1p2_m} shows these indices in the form of a table combining the human task conditions (LAD/HAD) and the models for these conditions (mLAD/mHAD).

Since higher values for R and lower values for RMSE imply a good fit, the corresponding combinations (LAD-mLAD and HAD-mHAD) are expected to be lighter in the heatmap of correlation coefficients, while those combinations are expected to be darker in the heatmap of RMSE. As a result, a good fit of LAD and mLAD was obtained for parameter 1 in both the correlation (0.8) and RMSE (1.7). This result shows that the low arousal model successfully simulated a decreasing trend in probe response time. The mechanism behind this learning is considered as the base-level learning, presented in equation \ref{baselevel_f}. As the number of presenting the probe ($n_{probe}$) increases, the difference in the activation values between the main goal ($A_{MainGoal}$) and the subgoal ($A_{SubGoal}$) decreases. 

On the other hand, for the combination of HAD and mHAD, it is difficult to find satisfactory fits between the human and simulation data. The correlations for this combination are small for both parameters (parameter 1: 0.066, parameter 2: 0.1). Also, as shown in Fig. \ref{fig:p1p2_m}, the RMSE for parameter 2 (1.4) is smaller than that for parameter 1 (7.4e+0.2). However, this value is larger than that for the combination of LAD and mHAD (1).

The reason of the low correlations between HAD and mHAD could be attributed to the lack of consistent temporal variation in the human probe response data.% as indicated in Section \ref{human_data}. 
As shown in Fig. \ref{fig:human}, the high arousal demands introduced large noises that were unrelated to the duration of the task. 
Therefore, in order to validate the model under the HAD condition, it is necessary to examine patterns in the data other than temporal variation. 

\begin{figure}[tb]
    \centering
    \includegraphics[width=84mm]{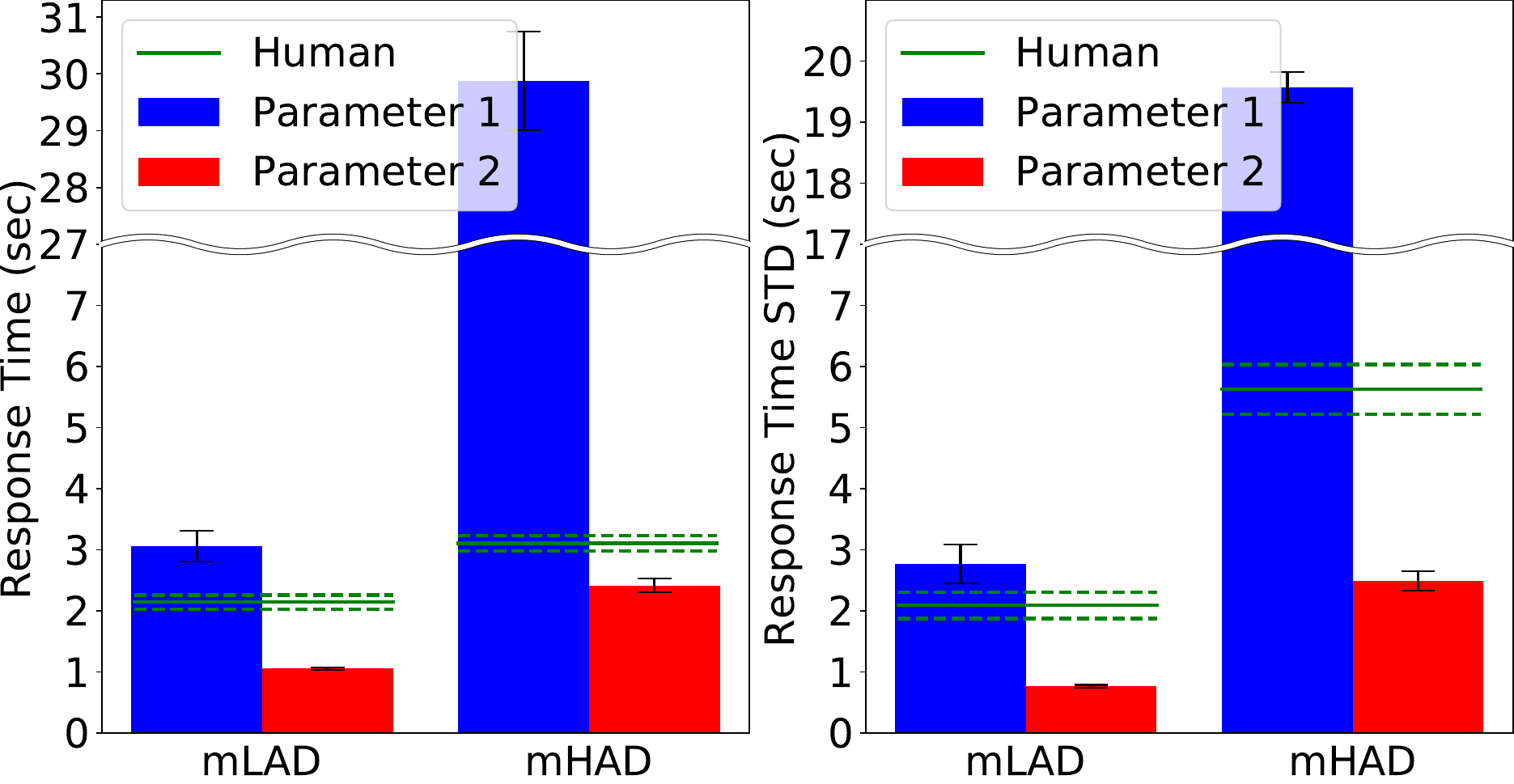}
 \caption{Mean and standard deviation of probe response time. The mean for each round shown in Figs \ref{fig:p1} and  \ref{fig:p2} were summed by condition and then averaged across 30 time points. Error bars represent a standard error of the mean  ($n=30$).}
 \label{fig:p1p2_b}
\end{figure}

Because of that, we calculated averages of the data points in the middle (averages over runs) and the right (STD over runs) panels of Figs \ref{fig:p1} and \ref{fig:p2}. Fig. \ref{fig:p1p2_b} summarizes the indices for each parameter and each model along with the human data (the green lines). As indicated in the figure, regardless of the difference of the parameter setting, high demand arousal led longer response time and higher deviation among runs as same as the human experiments. These results confirm that the mechanism of the high arousal presented in Section \ref{reward} could reproduce similar patterns observed in the human high-arousal demand condition.

\section{Conclusion}
\subsection{Summary and Implications}
This study aimed to develop a model of task immersion that leads to performance deficits caused by arousal dynamics. To achieve this goal, the ACT-R model of a multitasking environment was constructed following existing models of a perceptual-motor coordination task. Focusing on the overlooking subgoal-related stimuli, we set simulations under the low and high arousal demands (mLAD and mHAD) with different sets of parameters. The results showed a performance deficit in the subgoal due to the high arousal demand, similar to the corresponding human condition. Summarizing the results, our model successfully represents homeostatic regulations on arousal dynamics producing human compatible data. 

The significance of this study is the integrated account of the relationship between arousal and task performance based on the activation theory. Concerning the relationship between arousal and performance, the optimal level theory \cite{yerkes1908relation} has been used as a general explanation. Consistent with this theory, our ACT-R model showed the performance deficit caused by high arousal demands. We consider that this implemented mechanism is consistent with the cue utilization theory \cite{easterbrook1959effect}, which explains the narrowing attentional focus with increased arousal. 

Based on the discussion so far, we claim that using cognitive architecture to predict human arousal dynamics is advantageous to develop an integrated theory. There are previous studies dealing with the performance decrease at low arousal level \cite{nishikawa2022acs,nagashima2022iccm}. We believe that the results of the past studies will be integrated with those of the present study to provide a complete explanation of the organic arousal system. 

\subsection{Future Studies}

Although the above mentioned success and implications, the current study has a lot of limitations leading further research. 

The first one is in the human experiment, where participants were recruited using crowdsourcing. Due to the nature of online experiments, we could not obtain detailed data showing arousal dynamics caused by mismatches between actual and demanded arousal levels. Therefore, a lab experiment measuring physiological data, such as heart rate, is needed to validate the mechanism and the prediction of the model.

A related limitation is the manipulation of arousal demands, as this study primarily focused on task difficulty, which directly influences task performance. Therefore, it is hard to distinguish deficits caused by high arousal from performance declines simply due to the difficulty. 
To demonstrate the advantage of including arousal dynamics, we need to compare it to multitasking models without this factor, such as  \cite{kujala2015modeling}. 

Another important topic for future exploration is the representation of arousal. In the current research, we conceptualize arousal as an overall activation of the system based on its definition \cite{easterbrook1959effect}. This assumption leads to the inclusion of explicit goal retrieval within the model (see Fig. 4). Although the retrievals of chunks in this study does not introduce a time delay in the overall process because it has high activation from the beginning\footnote{The time required to complete the cycle is comparable to that of the previous model used in the line-following task \cite{morita2020cognitive}.}, declarative processes of ACT-R are generally considered to be costly and are gradually bypassed through learning \cite{kim2015learning}. The current study does not address the mechanism of attention allocation after such learning, where the model's process is primarily driven by the production rule selection. 

This limitation is also related to the probe confirmation process depicted in Fig. 4. In the current model, the probe is always detected as soon as it appears in the window. However, in deeply immersed situations, delays might occur before the probe information is stored in the goal buffer. To represent such delays, we could incorporate a mechanism of the selection between the confirm probe and the other productions relating the main goal. In ACT-R, such a conflict resolution between production rules is modulated by a parameter called utility. However, ACT-R does not currently address the relationship between the subsymbolic parameter for the production module and that for the declarative module. Therefore, future research is needed to explore these subsymbolic processes in ACT-R in a consistent manner to represent whole picture of the optimal arousal theory.
 
The final line of future study is exploring parameter fitting. Both parameters in Table 3 didn't lead to enough matching. Parameter 1 that focused more on the main goal  resulted in extremely high percentage of the missing response. Contrary, parameter 2 didn't reach at the same level of the response delay and deviation as the human data. These results suggest that the corresponding parameters for the human participants are somewhere between those parameter sets. In the future, we will adjust the parameter more precisely to explore the factors influencing arousal by instructions of the experiment.

\begin{table*}[tb]
\centering
\caption{Condition-action correspondence.}\label{table:actions}
\scalebox{0.55}[0.55]{
\begin{tabular}{c|cc|cc|cc|cc|cccccc}
\hline
                                                   & \multicolumn{2}{c|}{Left}                                                                                                                                                                                                                   & \multicolumn{2}{c|}{Left Punch}                                                                                                                                                                                                             & \multicolumn{2}{c|}{Right}                                                                                                                                                                                                                  & \multicolumn{2}{c|}{Right Punch}                                                                                                                                                                                                            & \multicolumn{6}{c}{Stop}                                                                                                                                                                                                                                                                                                                                                                                                                                                             \\ \hline
\begin{tabular}[c]{@{}c@{}}Current\\ Action\end{tabular} & \multicolumn{1}{c|}{Stop}                                                                                                                & Stop                                                                                             & \multicolumn{1}{c|}{Stop}                                                                                                                & Stop                                                                                             & \multicolumn{1}{c|}{Stop}                                                                                                                & Stop                                                                                             & \multicolumn{1}{c|}{Stop}                                                                                                                & Stop                                                                                             & \multicolumn{1}{c|}{Left}                                                                                   & \multicolumn{1}{c|}{Left}                                                                                 & \multicolumn{1}{c|}{Left} & \multicolumn{1}{c|}{Right}                                                                                  & \multicolumn{1}{c|}{Right}                                                                                & Right \\ \hline
Position                                                 & \multicolumn{1}{c|}{Right}                                                                                                               & Right                                                                                            & \multicolumn{1}{c|}{Right}                                                                                                               & Right                                                                                            & \multicolumn{1}{c|}{Left}                                                                                                                & Left                                                                                             & \multicolumn{1}{c|}{Left}                                                                                                                & Left                                                                                             & \multicolumn{1}{c|}{Right}                                                                                  & \multicolumn{1}{c|}{Right}                                                                                & \multicolumn{1}{c|}{Left} & \multicolumn{1}{c|}{Left}                                                                                   & \multicolumn{1}{c|}{Left}                                                                                 & Right \\ \hline
Goal                                                     & \multicolumn{1}{c|}{Far}                                                                                                                 & Near                                                                                             & \multicolumn{1}{c|}{Far}                                                                                                                 & Near                                                                                             & \multicolumn{1}{c|}{Far}                                                                                                                 & Near                                                                                             & \multicolumn{1}{c|}{Far}                                                                                                                 & Near                                                                                             & \multicolumn{1}{c|}{Far}                                                                                    & \multicolumn{1}{c|}{Near}                                                                                 & \multicolumn{1}{c|}{}     & \multicolumn{1}{c|}{Far}                                                                                    & \multicolumn{1}{c|}{Near}                                                                                 &       \\ \hline
Online                                                   & \multicolumn{1}{c|}{}                                                                                                                    & True                                                                                            & \multicolumn{1}{c|}{}                                                                                                                    &                                                                                             & \multicolumn{1}{c|}{}                                                                                                                    & True                                                                                            & \multicolumn{1}{c|}{}                                                                                                                    &                                                                                             & \multicolumn{1}{c|}{}                                                                                       & \multicolumn{1}{c|}{}                                                                                     & \multicolumn{1}{c|}{}     & \multicolumn{1}{c|}{}                                                                                       & \multicolumn{1}{c|}{}                                                                                     &       \\ \hline
Tracker                                                  & \multicolumn{1}{c|}{\begin{tabular}[c]{@{}c@{}}$vg_x \geqq vg_y + \theta_5$\\ $vg_x > \theta_1$\\ $vl \geqq vg + \theta_4$\end{tabular}} & \begin{tabular}[c]{@{}c@{}}$vl > \theta_1$\\ $vl > \theta_2$\\ $vl < vg + \theta_4$\end{tabular} & \multicolumn{1}{c|}{\begin{tabular}[c]{@{}c@{}}$vg_x < vg_y + \theta_5$\\ $vg_x > \theta_1$\\ $vl \geqq vg + \theta_4$\end{tabular}} & \begin{tabular}[c]{@{}c@{}}$vl > \theta_1$\\ $vl \leqq \theta_2$\\ $vl < vg + \theta_4$\end{tabular} & \multicolumn{1}{c|}{\begin{tabular}[c]{@{}c@{}}$vg_x \geqq vg_y + \theta_5$\\ $vg_x > \theta_1$\\ $vl \geqq vg + \theta_4$\end{tabular}} & \begin{tabular}[c]{@{}c@{}}$vl > \theta_1$\\ $vl > \theta_2$\\ $vl < vg + \theta_4$\end{tabular} & \multicolumn{1}{c|}{\begin{tabular}[c]{@{}c@{}}$vg_x < vg_y + \theta_5$\\ $vg_x > \theta_1$\\ $vl \geqq vg + \theta_4$\end{tabular}} & \begin{tabular}[c]{@{}c@{}}$vl > \theta_1$\\ $vl \leqq \theta_2$\\ $vl < vg + \theta_4$\end{tabular} & \multicolumn{1}{c|}{\begin{tabular}[c]{@{}c@{}}$vl \geqq vg + \theta_6$\\ $vg_x \leqq \theta_3$\end{tabular}} & \multicolumn{1}{c|}{\begin{tabular}[c]{@{}c@{}}$vl \leqq vg + \theta_6$\\ $vl \leqq \theta_3$\end{tabular}} & \multicolumn{1}{c|}{}     & \multicolumn{1}{c|}{\begin{tabular}[c]{@{}c@{}}$vl \geqq vg + \theta_6$\\ $vg_x \leqq \theta_3$\end{tabular}} & \multicolumn{1}{c|}{\begin{tabular}[c]{@{}c@{}}$vl \leqq vg + \theta_6$\\ $vl \leqq \theta_3$\end{tabular}} &       \\ \hline
\end{tabular}
}
\begin{tabular}{ccccccccccccccc}
\multicolumn{15}{l}{\parbox[t]{17cm}{* $vg$ denotes the distance to the goal. $vg_x$ and $vg_y$ denote its xy components. $vl$ denotes the distance to the line. Each $\theta$ is a correction value adjusted by the motor learning module.}}
\end{tabular}
\end{table*}
%\begin{acwwknowledgements}
%If you'd like to thank anyone, place your comments here
%and remove the percent signs.
%\end{acknowledgements}

% BibTeX users please use one of
\bibliographystyle{spbasic}      % basic style, author-year citations
% \bibliographystyle{spmpsci}      % mathematics and physical sciences
% \bibliographystyle{spphys}       % APS-like style for physics
% \bibliography{}   % name your BibTeX data base

\bibliography{test}
%
% and use \bibitem to create references. Consult the Instructions
% for authors for reference list style.
%
% \bibitem{RefJ}
% % Format for Journal Reference
% Author, Article title, Journal, Volume, page numbers (year)
% % Format for books
% \bibitem{RefB}
% Author, Book title, page numbers. Publisher, place (year)
% % etc
% \end{thebibliography}

\section*{Appendix}
\subsection*{Mastering Motor Control}
The accuracy of the perceptual-motor loop (Fig. \ref{fig:state}) is improved by learning through the task. This learning is controlled by a tracker module included in ACT-R 7.27, initially proposed by Anderson et al. \cite{anderson2019learning}. This module adjusts the continuous conditions for selecting motor operations based on positive and negative feedback from the environment and simulated annealing algorithm \cite{kirkpatrick1983optimization}.

As indicated in Section \ref{motor_module}, the model performs six key operations. These behaviors are determined by the value of the correction value related to the continuous conditions. The module learns those values by evaluating them against the results of the model's actions.

In this study, positive feedback is generated when the model goes from the offline state to the online state and the model is close to the line during the offline state. Negative feedback is generated in the opposite pattern. Based on the values of these feedbacks, the parameter $\theta$ in Table \ref{table:actions} is adjusted to be optimal to follow the line. The role of each row indicates as follows.
 
\begin{enumerate}
    \item {\it Current Action}: Current action.
    \item {\it Position}: Relative position of the model to the scrolling line.
    \item {\it Goal}: The goal that the model is heading. {\it Far} represents the subgoal and {\it Near} represents the scrolling line.
    \item {\it Online}: The distinction of the state that model is following the line.
    \item {\it Tracker}: 
Conflict resolution criteria using the vehicle coordinates and the motor learning module adjustment values.
\end{enumerate}

If this optimization fails, the model cannot trace the line because the vehicle either crosses the line or stops moving before reaching the line.

\end{document}